\documentclass[10pt,twocolumn,letterpaper]{article}
\usepackage[svgnames,table]{xcolor}

\usepackage{cvpr}
\usepackage{times}
\usepackage{epsfig}
\usepackage{graphicx}
\usepackage{amsmath}
\usepackage{amssymb}
\usepackage{listings}
\usepackage{makecell}
\usepackage{booktabs}
\usepackage{multirow}
\usepackage{color}
\usepackage{comment}

\usepackage[pagebackref=true,breaklinks=true,letterpaper=true,colorlinks,bookmarks=false]{hyperref}

\cvprfinalcopy %

\def\cvprPaperID{358} %

\def\Att{{V_{att}}}
\def\Cap{{V_{cap}}}
\def\Know{{V_{know}}}
\def\ModParms{{\boldsymbol \theta}}

\ifcvprfinal\fi
\begin{document}

\title{Ask Me Anything: Free-form Visual Question Answering \\Based on Knowledge from External Sources}

\author{Qi Wu, Peng Wang, Chunhua Shen, Anthony Dick, Anton van den Hengel\\
School of Computer Science, 
The University of Adelaide, Australia\\
{\tt\small \{qi.wu01,p.wang,chunhua.shen,anthony.dick,anton.vandenhengel\}@adelaide.edu.au}}

\maketitle
\thispagestyle{empty}

\begin{abstract}

We propose a method for visual question answering which combines an internal representation of the content of an image with information extracted from a general knowledge base to answer a broad range of image-based questions. This allows more complex questions to be answered using the predominant neural network-based approach than has previously been possible.  It particularly allows questions to be asked about the contents of an image, even when the image itself does not contain the whole answer. The method constructs a textual representation of the semantic content of an image, and merges it with textual information sourced from a knowledge base, to develop a deeper understanding of the scene viewed. Priming a recurrent neural network with this combined information, and the submitted question, leads to a very flexible visual question answering approach.   We are specifically able to answer questions posed in natural language, that refer to information not contained in the image.
We demonstrate the effectiveness of our model on two publicly available datasets, Toronto~COCO-QA \cite{ren2015image} and VQA~\cite{antol2015vqa} and show that it produces the best reported results in both cases.
\end{abstract}

\section{Introduction}

Visual question answering (VQA) is distinct from many problems in Computer Vision because the question to be answered is not determined until run time~\cite{antol2015vqa}.  In more traditional problems such as segmentation or detection, the single question to be answered by an algorithm is predetermined, and only the image changes.  In visual question answering, in contrast, the form that the question will take is unknown, as is the set of operations required to answer it.  In this sense it more closely reflects the challenge of general image interpretation.

VQA typically requires processing both visual information (the image) and textual information (the question and answer).  One approach to Vision-to-Language problems, such as VQA and image captioning, which interrelate visual and textual information is based on a direct method pioneered in machine language translation~\cite{cho2014learning,sutskever2014sequence}. 
This direct approach develops an encoding of the input text using a Recurrent Neural Network (RNN) and passes it to another RNN for decoding. %

\begin{figure}[t]
  \includegraphics[width=1\linewidth]{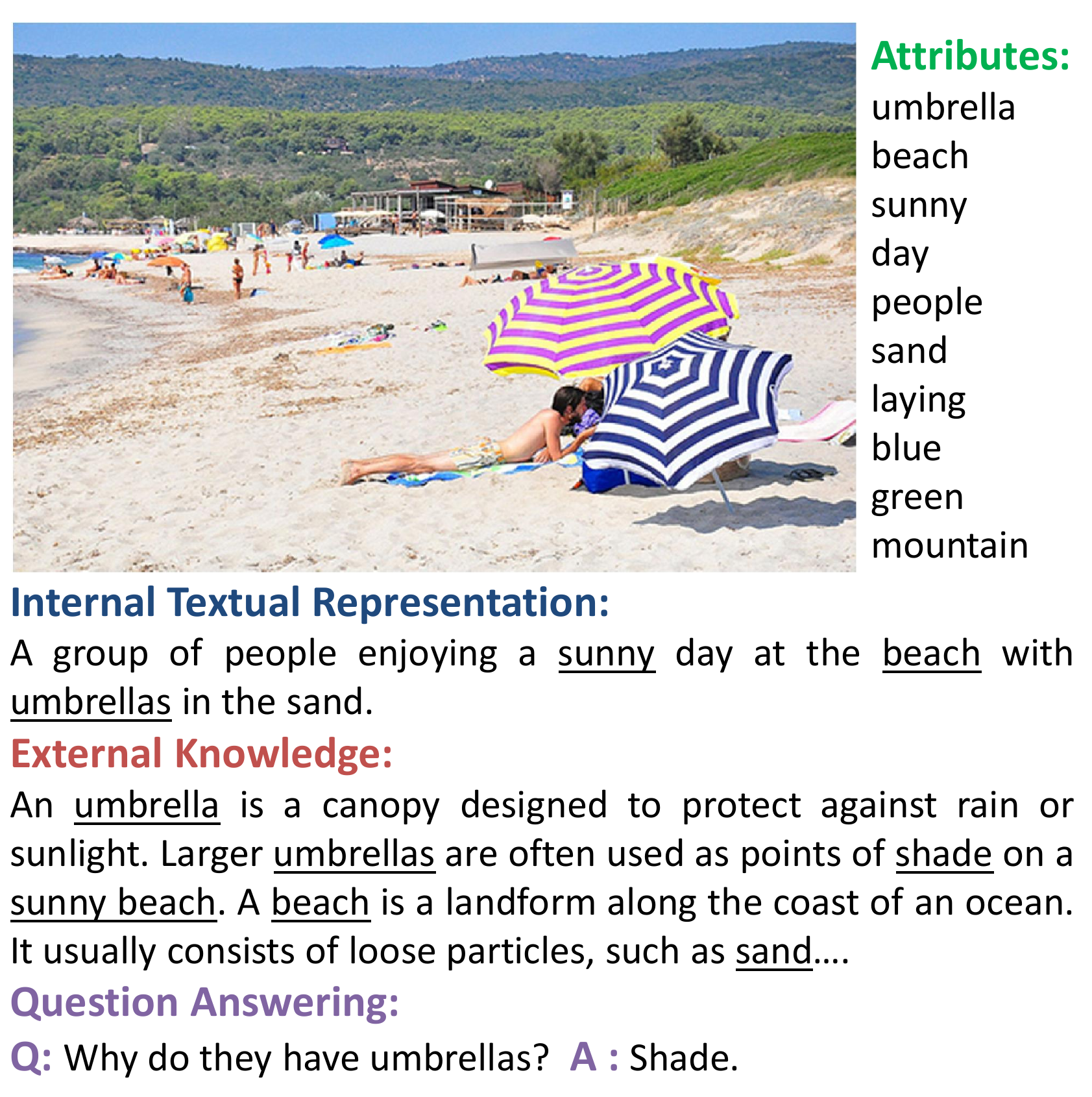}
  \caption{A real case of question answering based on an internal textual representation and external knowledge. All of the attributes, textual representation, knowledge and answer are produced by our VQA model. Underlined words indicate the information required to answer the question.}
  \label{img:example}
\end{figure}

The Vision-to-Language version of this direct translation approach uses a Convolutional Neural Network (CNN) to encode the image, and an RNN to encode and generate text.
This approach has been well studied.  The image captioning methods developed in~\cite{donahue2014long,karpathy2014deep,mao2014deep,vinyals2014show,yao2015describing}, for example, 
learn a mapping from images to text 
using CNNs trained on object recognition, and word embeddings trained on large scale text corpora. 
Visual question answering is a significantly more complex problem than image captioning, not least because it requires accessing information not present in the image.  This may be common sense, or specific knowledge about the image subject.
For example, given an image, such as Figure \ref{img:example}, showing `a group of people enjoying a sunny day at the beach with umbrellas', if one asks a question `why do they have umbrellas?', to answer this question, the machine must not only detect the scene `beach', but must know that `umbrellas are often used as points of shade on a sunny beach'. Recently, Antol \etal~\cite{antol2015vqa} also have suggested that VQA is a more ``AI-complete'' task since it requires multimodal knowledge beyond a single sub-domain. Our proposed system finally gives the right answer `shade' for the above real example.

Large-scale Knowledge Bases (KBs), such as Freebase~\cite{bollacker2008freebase} and DBpedia~\cite{auer2007dbpedia}, have been used successfully in several
natural language Question Answering (QA) systems~\cite{berant2013semantic,ferrucci2010building}.
 However, VQA systems exploiting KBs 
are still relatively rare.
Gao \etal~\cite{gao2015you} and Malinowski \etal~\cite{malinowski2015ask} 
do not use a KB at all.
Zhu \etal \cite{zhu2015building} do use a KB, but it is created specifically for the purpose, and consequently contains a small amount of very specific, largely image-focused, information.  This in turn means that only a specific set of questions may be asked of the system, and the query generation approach taken mandates a very particular question form.  The method that we propose here, in contrast, is applicable to general (even publicly created)  KBs, and admits general questions.  

\begin{figure*}[t]
\begin{center}
\includegraphics[width=0.99\linewidth]{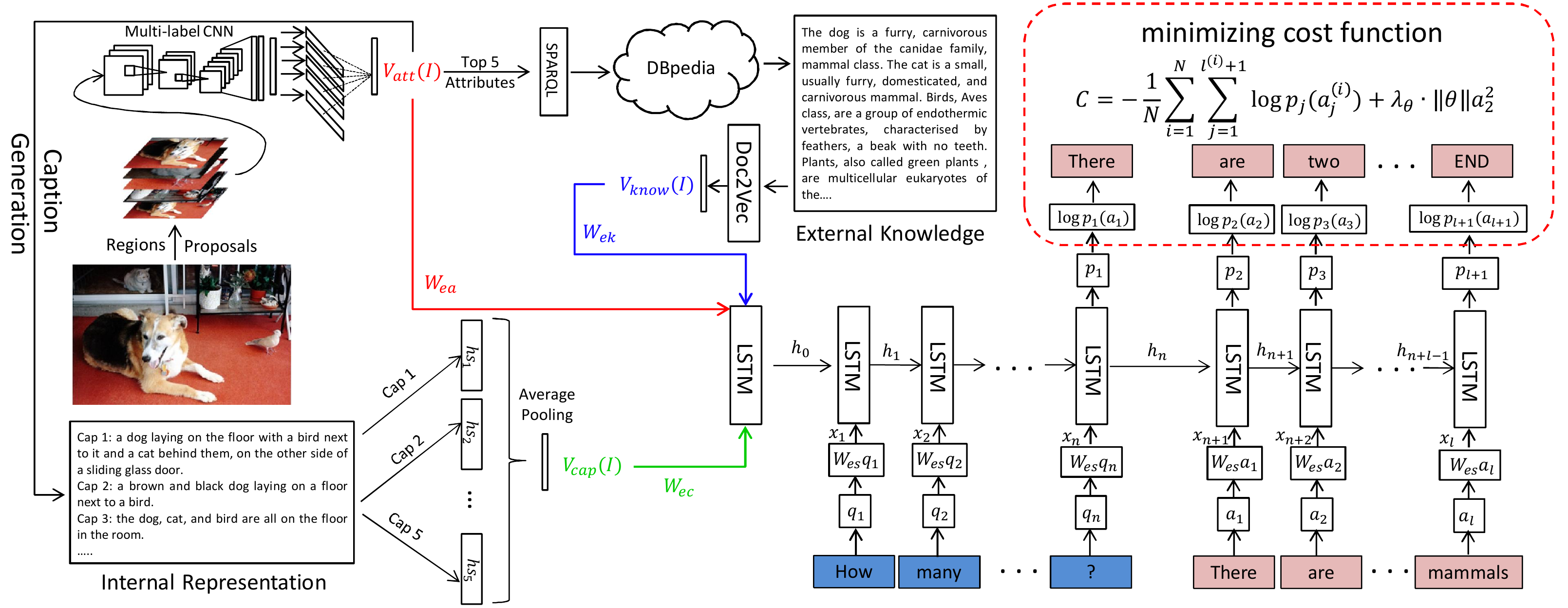}
  \vspace{-1pt}
  \caption{Our proposed framework: given an image, a CNN is first applied to produce the attribute-based representation \textcolor{red}{$\Att(I)$}. The internal textual representation is made up of image captions generated based on the image-attributes. The hidden state of the caption-LSTM after it has generated the last word in each caption is used as its vector representation. These vectors are then aggregated as \textcolor{green}{$\Cap(I)$} with average-pooling. The external knowledge is mined from the KB (in this case DBpedia) and the responses encoded by Doc2Vec, which produces a vector \textcolor{blue}{$\Know(I)$}. The 3 vectors $\mathbf{V}$ are combined into a single representation of scene content,  which is input to the VQA LSTM model which interprets the question and generates an answer.}
\end{center}
\label{frame}
\vspace{-10pt}
\end{figure*}

In this work, we 
fuse an automatically generated description of an image with information extracted from an external KB to provide an answer to a general question about the image (See Figure \ref{frame}). The image description takes the form of a set of captions, and the external knowledge is text-based information mined from a Knowledge Base. Given an image-question pair, a CNN is first employed to predict a set of attributes of the image. The attributes cover a wide range of high-level concepts, including objects, scenes, actions, modifiers and so on.  A state-of-the-art image captioning model \cite{wu2015image} is applied to generate a series of captions based on the attributes. 
We then use the detected attributes to extract relevant information from the KB.  Specifically, for each of the top-5 attributes detected in the image we generate a query which may be applied to a Resource Description Framework (RDF) KB, such as DBpedia.  RDF is the standard format for large KBs, of which there are many.  The queries are specified using Semantic Protocol And RDF Query Language (SPARQL). 
We encode the paragraphs extracted from the KB using Doc2Vec \cite{le2014distributed}, which maps paragraphs into a fixed-length feature representation. 
The encoded attributes, captions, and KB information are then input to an LSTM which is trained so as to maximise the likelihood of the ground truth answers in a training set.

The approach we propose here combines the generality of information that using a KB allows with the generality of question that the LSTM allows.
In addition, it achieves an accuracy of 69.73\% on the Toronto COCO-QA, while the latest state-of-the-art is 55.92\%. %
We also produce the best results on the VQA evaluation server (which does not publish ground truth answers for its test set), which is 59.44\%.
\section{Related Work}
Malinowski \etal \cite{malinowski2014multi} were among the first to study the VQA problem. They proposed a method that combines semantic parsing and image segmentation with a Bayesian approach to sample from nearest neighbors in the training set. This approach requires human defined predicates, which are inevitably dataset-specific. This approach is also very dependent on the accuracy of the image segmentation algorithm and on the estimated image depth information. Tu \etal~\cite{tu2014joint} built a query answering system based on a joint parse graph from text and videos. Geman \etal~\cite{geman2015visual} proposed an automatic `query generator' that was trained on annotated images and produced a sequence of binary questions from any given test image. 
Each of these approaches places significant limitations on the form of question that can be answered.

Most recently, inspired by the significant progress achieved using deep neural network models in both computer vision and natural language processing, an architecture which combines a CNN and RNN to learn the mapping from images to sentences has become the dominant trend. Both Gao \etal~\cite{gao2015you} and Malinowski \etal \cite{malinowski2015ask} used RNNs to encode the question and output the answer.  Whereas Gao \etal~\cite{gao2015you} used two networks, a separate encoder and decoder, Malinowski \etal \cite{malinowski2015ask} used a single network for both encoding and decoding. Ren \etal \cite{ren2015image} focused on questions with a single-word answer and formulated the task as a classification problem using an LSTM. A single-word answer dataset COCO-QA was published with \cite{ren2015image}. Ma \etal \cite{ma2015learning} used CNNs to both extract image features and sentence features, and fused the features together with another multimodal CNN. Antol \etal \cite{antol2015vqa} proposed a large-scale open-ended VQA dataset based on COCO, which is called VQA. They also provided several baseline methods which combined both image features (CNN extracted) and question features (LSTM extracted) to obtain a single embedding and further built a MLP (Multi-Layer Perceptron) to obtain a distribution over answers. Our framework also exploits both CNNs and RNNs, but in contrast to preceding approaches which use only image features extracted from a CNN in answering a question, we employ multiple sources, including image content, 
generated image captions and mined external knowledge, to feed to an RNN to answer questions.

The quality of the information in the KB is one of the primary issues in this approach to VQA. The problem is that KBs constructed by analysing Wikipedia and similar are patchy and inconsistent at best, and hand-curated KBs are inevitably very topic specific. Using visually-sourced information is a promising approach to solving this problem~\cite{Lin_2015_CVPR, Sadeghi_2015_CVPR}, but has a way to go before it might be usefully applied within our approach.  Thus, although our SPARQL and RDF driven approach can incorporate any information that might be extracted from a KB, the limitations of the existing available KBs mean that the text descriptions of the detected attributes is all that can be usefully extracted.

Zhu \etal \cite{zhu2015building}, in contrast used a hand-crafted KB primarily containing image-related information such as category labels, attribute labels and affordance labels, but also some quantities relating to their specific question format such as GPS coordinates and similar. 
The questions in that system are phrased in the DBMS query language, and are thus tightly coupled to the nature of the hand-crafted KB.  This represents a significant restriction on the form of question that might be asked, but has the significant advantage that the DBMS is able to respond decisively as to whether it has the information required to answer the question. Instead of building a problem-specific KB, we use a pre-built large-scale KB (DBpedia \cite{auer2007dbpedia}) from which we extract information using a standard RDF query language. DBpedia has been created by extracting structured information from Wikipedia, and is thus significantly larger and more general than a hand-crafted KB. Rather than having a user to pose their question in a formal query language, our VQA system is able to encode questions written in natural language automatically.  This is achieved without manually specified formalization, but rather depends on processing a suitable training set. The result is a model which is very general in the forms of question that it will accept.
\section{Extracting, Encoding, and Merging}
\label{Sec:Feature_Extract}
The key differentiator of our approach is that it is able to usefully combine image information with that extracted from a KB, within the LSTM framework.  The novelty lies in the fact that this is achieved by representing both of these disparate forms of information as text before combining them.
Figure~\ref{frame} summarises how this is achieved.
We now describe each step in more details. 

\subsection{Attribute-based Image Representation}
Our first task is to describe the image content in terms of a set of attributes. 
Each attribute in our vocabulary is extracted from captions from MS COCO \cite{chen2015microsoft}, a large-scale image-captioning dataset. An attribute can be any part of speech, including object names (nouns), motions (verbs) or properties (adjectives). Because MS COCO captions are created by humans, attributes derived from captions 
are likely to represent image features that are of particular significance to humans, and are thus likely to be the subject of image-based questions.

We formalize attribute prediction as a multi-label classification problem. To address the issue that some attributes may only apply to image sub-regions, we follow Wei \etal~\cite{wei2014cnn} to design a region-based multi-label classification framework that takes an arbitrary number of sub-region proposals as input. A shared CNN is connected with each proposal, and the CNN outputs from different proposals are aggregated with max pooling to produce the final prediction over the attribute vocabulary. 

To initialize the attribute prediction model, we use the powerful VggNet-16 \cite{simonyan2014very} pre-trained on the ImageNet~\cite{deng2009imagenet}. The shared CNN is then fine-tuned on the multi-label dataset, the MS COCO image-attribute training data~\cite{wu2015image}. The output of the last fully-connected layer is fed into a $c$-way softmax which produces a probability distribution over the $c$ class labels. The $c$ represents the attribute vocabulary size, which here is 256. The fine-tuning learning rates of the last two fully connect layers are initialized to 0.001 and the prediction layer is initialized to 0.01. All the other layers are fixed. We execute 40 epochs in total and decrease the learning rate to one tenth of the current rate for each layer after 10 epochs. The momentum is set to 0.9. The dropout rate is set to 0.5. Then, we use Multiscale Combinatorial Grouping (MCG)~\cite{PABMM2015} for the proposal generation.
Finally, a cross hypothesis max-pooling is applied to integrate the outputs into a single prediction vector $\Att(I)$.

\begin{figure}[t]
\begin{center}
   \includegraphics[width=1\linewidth]{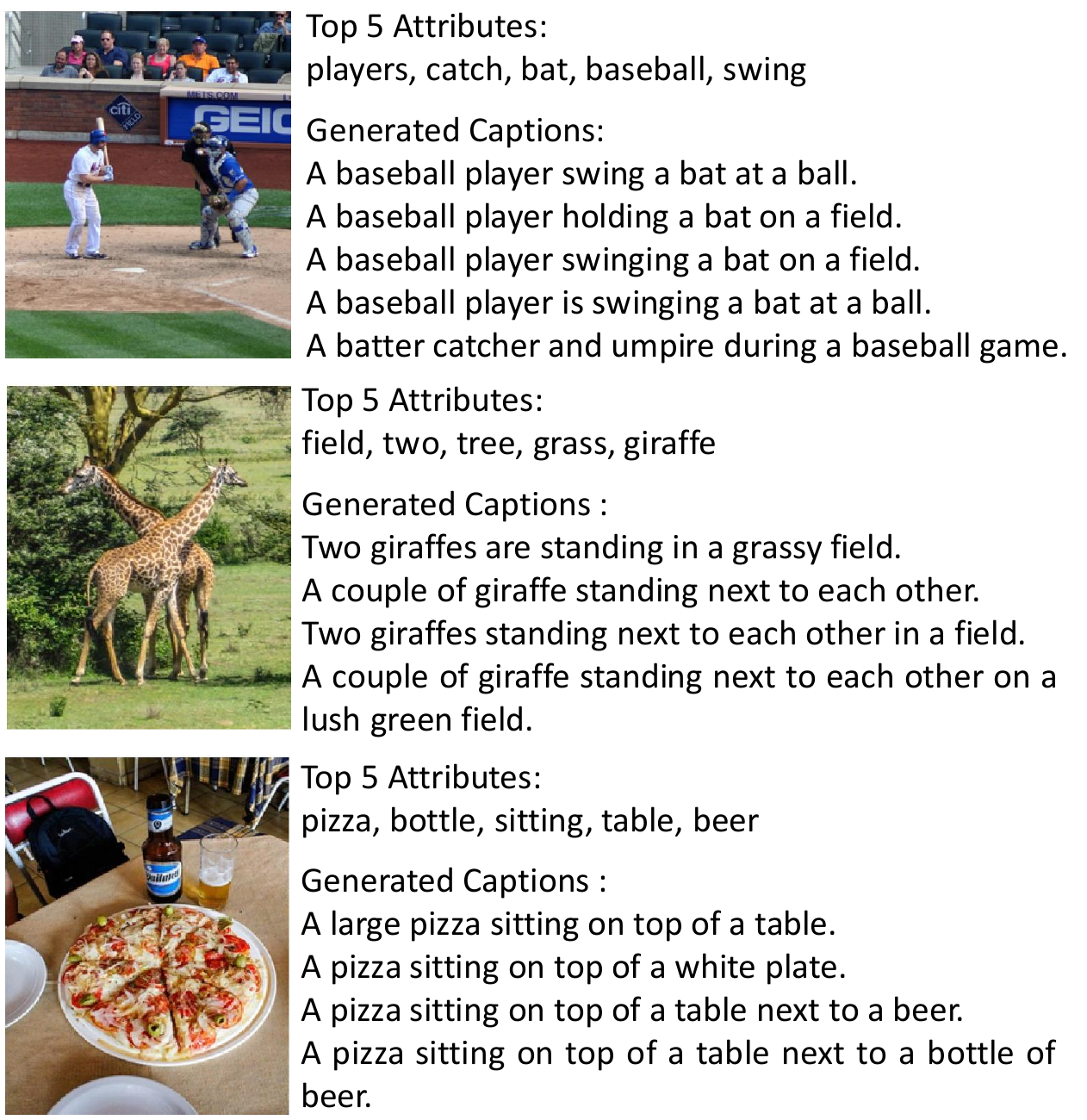}
\end{center}
\vspace{-10pt}
   \caption{Examples of predicted attributes and generated captions.}
   \vspace{-5pt}
   \label{att_caP_examples}
\end{figure}

\subsection{Caption-based Image Representation}
\label{subsec:Story}
Currently the most successful approach to image captioning ~\cite{chen2014learning,donahue2014long,karpathy2014deep,mao2014deep,vinyals2014show} is to attach a CNN to an RNN to learn the mapping from images to sentences directly. Wu \etal~\cite{wu2015image} proposed to feed a high-level attribute-based representation to an LSTM to generate captions, instead of directly using  CNN-extracted features. This method produces promising results on the major public captioning challenge~\cite{chen2015microsoft} and accepts our attributes prediction vector $\Att(I)$ as the input. We thus use this approach to generate 5 different captions (using beam search) that constitute the internal textual representation for a given image. %

The hidden state vector of the caption-LSTM after it has generated the last word in each caption is used to represent its content. %
Average-pooling is applied over the 5 hidden-state vectors, to obtain a 512-d vector $\Cap(I)$ for the image~$I$. The caption-LSTM is trained on the human-generated captions from the MS COCO training set, which means that the resulting model is focused on the types of image content that humans are most interested in describing. Figure~\ref{att_caP_examples} shows some examples of the predicted attributes and generated captions.

\subsection{Relating to the Knowledge Base}
\begin{figure}[h]
\vspace{-5pt}
  \centering
  \includegraphics[width=0.9\linewidth]{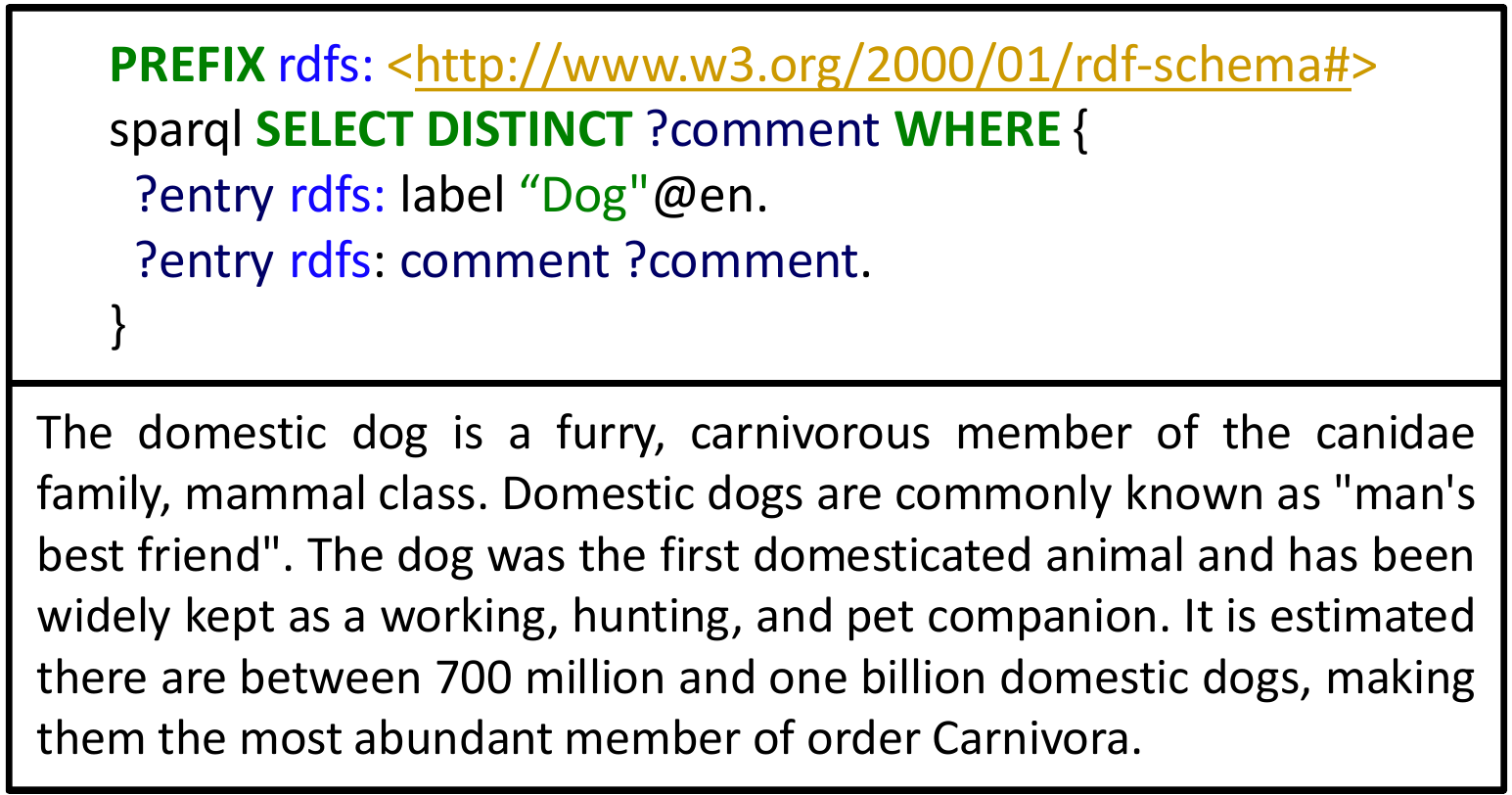}
  \caption{An example of SPARQL query language for the attribute \texttt{`dog'}. The mined text-based knowledge are shown below.}
  \label{img:example_rdf}
  \vspace{-8pt}
\end{figure}

The external data source that we use here is
DBpedia~\cite{auer2007dbpedia}. As a source of general background information, although any such KB could equally be applied, DBpedia is a structured database of information extracted from Wikipedia. The whole DBpedia dataset describes $4.58$ million entities, of which $4.22$ million are classified in a consistent ontology. The data can be accessed using an SQL-like query language for RDF called SPARQL. Given an image and its predicted attributes, we use the top-5 most strongly predicted attributes\footnote{We only use the top-5 attributes to query the KB because, based on the observation of training data, an image typically contains 5-8 attributes. We also tested with top-10, but no improvements were observed.} to generate DBpedia queries. 
Inspecting the database shows that the `comment' field is the most generally informative about an attribute, as it contains a general text description of it.  We therefore retrieve the comment text for each query term. The KB+SPARQL combination is very general, however, and could be applied to problem specific KBs, or a database of common sense information, and can even perform basic inference over RDF. Figure \ref{img:example_rdf} shows an example of the query language and returned text.

Since the 
text returned by the SPARQL query is typically 
much longer than the captions generated in the section \ref{subsec:Story}, we turn to Doc2Vec \cite{le2014distributed} to extract the semantic meanings. %
Doc2Vec, also known as Paragraph Vector, is an unsupervised algorithm that learns fixed-length feature representations from variable-length pieces of texts, such as sentences, paragraphs, and documents. Le \etal \cite{le2014distributed}  proved that it can capture the semantics of paragraphs. A Doc2Vec model is trained to predict words in the document given the context words. We collect 100,000 documents from  DBpedia to train a model with vector size 500. To obtain the knowledge vector $\Know(I)$ for image $I$, we combine the 5 returned paragraphs in to a single large paragraph, before extracting semantic features using our Doc2Vec model.

\section{A VQA Model with Multiple Inputs}
\vspace{-2pt}
\label{Sec:Answer_Generation}
We propose to train a VQA model by maximizing the probability of the correct answer given the image and question. 
We want our VQA model to be able to generate multiple word answers, so we formulate the answering process as a word sequence generation procedure. Let $Q=\{q_1,...,q_n\}$ represents the sequence of words in a question, and $A=\{a_1,...,a_l\}$ the answer sequence, where $n$ and $l$ are the length of question and answer, respectively. The log-likelihood of the generated answer can be written as: 
\vspace{-7pt}
\begin{equation}
    \log p(A|I,Q)=\sum_{t=1}^l \log p(a_{t}|a_{1:t-1},I,Q)
\vspace{-3pt}    
\end{equation}
where $p(a_t|a_{1:t-1},I,Q)$ is the probability of generating $a_t$ given image information $I$, question $Q$ and previous words $a_{1:t-1}$. We employ an encoder LSTM~\cite{hochreiter1997long} to take the semantic information from image $I$ and the question $Q$, while using a decoder LSTM to generate the answer. Weights are shared between the encoder and decoder LSTM.

In the training phase, the question $Q$ and answer $A$ are concatenated as $\{q_1,...,q_n,a_1,...,a_l,a_{l+1}\}$, where $a_{l+1}$ is a special END token. Each word is represented as a one-hot vector of dimension equal to the size of the word dictionary. The training procedure is as follows: at time step $t=0$, we set the LSTM input: 
\vspace{-3pt}
\begin{equation}
   x_{initial}=[ W_{ea}\Att(I), W_{ec}\Cap(I), W_{ek}\Know(I) ]
\vspace{-2pt}   
\end{equation}
where $W_{ea}$, $W_{ec}$, $W_{ek}$ are learnable embedding weights for the vector representation of attributes, captions and external knowledge, respectively. In practice, all these embedding weights are learned jointly. Given the randomly initialized hidden state, the encoder LSTM feeds forward to produce  hidden state $h_{0}$ which encodes all of the input information. From $t=1$ to $t=n$, we set $x_t=W_{es}q_t$ and the hidden state $h_{t-1}$ is given by the previous step, where $W_{es}$ is the learnable word embedding weights. The decoder LSTM runs from time step $n+1$ to $l+1$. Specifically, at time step $t=n+1$, the LSTM layer takes the input $x_{n+1}=W_{es}a_1$ and the hidden state $h_{n}$ corresponding to the last word of the question, where $a_1$ is the start word of the answer. The hidden state $h_{n}$ thus encodes all available information about the image and the question. The probability distribution $p_{t+1}$ over all answer words in the vocabulary is then computed by the LSTM feed-forward process. Finally, for the final step, when $a_{l+1}$ represents the last word of the answer, the target label is set to the END token.

Our training objective is to learn parameters $W_{ea}$, $W_{ec}$, $W_{ek}$, $W_{es}$ and all the parameters in the LSTM by minimizing the following cost function:
\vspace{-5pt}
\begin{eqnarray}
    \mathcal{C}&=&-\frac{1}{N}\sum_{i=1}^N\log p(A^{(i)}|I,Q)+\lambda_\ModParms\cdot||\ModParms||_2^2 \\
    &=&-\frac{1}{N}\sum_{i=1}^N\sum_{j=1}^{l^{(i)}+1}\log p_j(a_j^{(i)})+\lambda_\ModParms\cdot||\ModParms||_2^2
\vspace{-7pt}
\end{eqnarray}
where $N$ is the number of training examples, and $n^{(i)}$ and $l^{(i)}$ are the length of question and answer respectively for the $i$-th training example. Let $p_t(a_t^{(i)})$ correspond to the activation of the Softmax layer in the LSTM model for the $i$-th input and $\ModParms$ represent the model parameters. Note that $\lambda_\ModParms\cdot||\ModParms||_2^2$ is a regularization term. %
\section{Experiments}
\label{Sec:Experiment}
We evaluate our model on two recent publicly available visual question answering datasets, both based on MS COCO images. The Toronto COCO-QA Dataset \cite{ren2015image} contains 78,736 training and 38,948 testing examples, which are generated from 117,684 images. There are four types of questions, relating to the object, number, color and location, all constructed so as to have a single-word answer. All of the question-answer pairs in this dataset are automatically converted from human-sourced image descriptions. Another benchmarked dataset is VQA \cite{antol2015vqa}, which is a much larger dataset and contains 614,163 questions and 6,141,630 answers based on 204,721 MS COCO images. This dataset provides a surprising variety of question types, including ``What is...', ``How Many'' and even ``Why...''. The ground truth answers were generated by 10 human subjects and can be single word or sentences. The data train/val split follows the COCO official split, which contains 82,783 training images and 40,504 validation images, each has 3 questions and 10 answers. We randomly choose 5000 images from the validation set as our val set, 
with the remainder
testing. The human ground truth answers for the actual VQA test split are not available publicly and only can be evaluated via the VQA evaluation server. Hence, we also apply our final model on a test split and report the overall accuracy. Table \ref{tab1} displays some dataset statistics.
We did not test on the DAQUAR dataset~\cite{malinowski2014towards} as it is an order of magnitude smaller than the datasets mentioned above, and thus too small to train our system, and to test its generality.
\subsection{Implementation Details}
To train the VQA model with multiple inputs in the Section \ref{Sec:Answer_Generation}, we use Stochastic gradient Descent (SGD) with mini-batches of 100 image-QA pairs. The attributes, internal textual representation, external knowledge embedding size, word embedding size and hidden state size are all 256 in all experiments. The learning rate is set to 0.001 and clip gradients is 5. The dropout rate is set to 0.5.

\begin{table}[t]
\begin{center}
\scriptsize
\resizebox{\linewidth}{!}{
 \begin{tabular}{l c c}
 \Xhline{2\arrayrulewidth}
 &Toronto COCO-QA&VQA \\ \hline
 \# Images&117,684&204,721 \\
 \# Questions&117,684&614,163 \\
 \# Question Types&4&more than 20 \\
 \# Ans per Que&1&10\\
 \# Words per Ans&1&1 or multiple\\ \Xhline{2\arrayrulewidth}
 \end{tabular}}
 \vspace{1pt}
 \caption{Some statistics about the Toronto COCO-QA Dataset \cite{ren2015image} and VQA dataset \cite{antol2015vqa}.}
 \label{tab1}
 \vspace{-16pt}
\end{center}
\end{table}

\subsection{Results on Toronto COCO-QA}
\paragraph{Metrics}
Following \cite{ma2015learning, ren2015image}, the accuracy value (the proportion of correctly answered test questions),
and the Wu-Palmer similarity (WUPS) \cite{wu1994verbs} are used to measure performance. The WUPS calculates the similarity between two words based on the similarity between their common subsequence in the taxonomy tree. If the similarity between two words is greater than a threshold then the candidate answer is considered to be right. 
We report on thresholds 0.9 and 0.0, following~\cite{ma2015learning,ren2015image}.

\begin{table}[b!]
\vspace{-8pt}
\scriptsize
\begin{center}
\resizebox{\linewidth}{!}{
    \begin{tabular}{ l|c c c }
    \Xhline{2\arrayrulewidth}
    \textbf{Toronto COCO-QA} &Acc(\%) & WUPS@0.9 & WUPS@0.0 \\ \hline
    GUESS\cite{ren2015image}&6.65&17.42&73.44 \\
    VIS+BOW\cite{ren2015image}&55.92&66.78&88.99 \\
    VIS+LSTM\cite{ren2015image}&53.31&63.91&88.25 \\
    2-VIS+BLSTM\cite{ren2015image}&55.09&65.34&88.64 \\
    Ma et al.\cite{ma2015learning}&54.94&65.36&88.58\\
    \Xhline{2\arrayrulewidth}
    \textbf{Baseline} &  &   &   \\ \hline
    VggNet-LSTM & 50.73 & 60.37 & 87.48 \\
    VggNet+ft-LSTM & 58.34& 67.32 & 89.13 \\
    \Xhline{2\arrayrulewidth}
    \textbf{Our-Proposal} &  &   &   \\ \hline
    Att-LSTM& 61.38& 71.15& 91.58 \\
    Att+Cap-LSTM& 69.02& 76.20& 92.38 \\
    Att+Know-LSTM& 63.07& 72.22& 90.84 \\
    Cap+Know-LSTM& 64.31& 73.31& 90.01 \\
    Att+Cap+Know-LSTM& \textbf{69.73}& \textbf{77.14}& \textbf{92.50} \\
     \Xhline{2\arrayrulewidth}
    \end{tabular}}
    \vspace{1pt}
    \caption{Accuracy, WUPS metrics compared to other state-of-the-art methods and our baseline on Toronto COCO-QA dataset.}
    \label{tab2}
    \vspace{-10pt}
\end{center}
\end{table}

\vspace{-12pt}
\paragraph{Evaluations}
To illustrate the effectiveness of our model, we provide a baseline and several state-of-the-art results on the Toronto COCO-QA dataset. The \textbf{Baseline} method is implemented simply by connecting a CNN to an LSTM. The CNN is a pre-trained (on ImageNet) VggNet model 
from which we extract the coefficients of the last fully connected layer.
We also implement a baseline model \textbf{VggNet+ft-LSTM}, which applies a VggNet that has been fine-tuned on the COCO dataset, based on the task of image-attributes classification.
We also present results from a series of cut down versions of our approach for comparison.
\textbf{Att-LSTM} uses only the semantic level attribute representation $\Att$ as the LSTM input. To evaluate the contribution of the internal textual representation and external knowledge for the question answering, we feed the image caption representation $\Cap$ and knowledge representation $\Know$ with the $\Att$ separately, producing two models, \textbf{Att+Cap-LSTM} and \textbf{Att+Know-LSTM}. We also tested the \textbf{Cap+Know-LSTM}, for the experiment completeness. Our final model is the \textbf{Att+Cap+Know-LSTM}, which combines all the available information.

\textbf{GUESS} \cite{ren2015image} simply selects the modal answer from the training set for each of 4 question types (the modal answers are `cat', `two', `white', and `room'). 
\textbf{VIS+BOW}~\cite{ren2015image} performs multinomial logistic regression based on image features and a BOW vector obtained by summing all the word vectors of the question. \textbf{VIS+LSTM}~\cite{ren2015image} has one LSTM to encode the image and question, while \textbf{2-VIS+BLSTM}~\cite{ren2015image} has two image feature inputs, at the start and the end. Ma \textit{et al.}~\cite{ma2015learning} encode both images and questions with a CNN.

\begin{table}[t!]
\scriptsize
\resizebox{\linewidth}{!}{
    \begin{tabular}{ l|c c c c }
    \Xhline{2\arrayrulewidth}
    \textbf{COCO-QA} &Object & Number & Color & Location \\ \hline
    GUESS&2.11&35.84&13.87& 8.93\\
    VIS+BOW&58.66&44.10&51.96&49.39 \\
    VIS+LSTM&56.53&46.10&45.87&45.52 \\
    2-VIS+BLSTM&58.17&44.79&49.53&47.34 \\
    \Xhline{2\arrayrulewidth}
    \textbf{Baseline} & &   & &  \\ \hline
    VggNet-LSTM &53.71&45.37&36.23&46.37 \\
    VggNet+ft-LSTM &61.67&50.04&52.16&54.40 \\
    \Xhline{2\arrayrulewidth}
    \textbf{Our-Proposal} &  &   & &  \\ \hline
    Att-LSTM&63.92&51.83&57.29&54.84 \\
    Att+Cap-LSTM&71.30&69.98&61.50&\textbf{60.98} \\
    Att+Know-LSTM&64.57&54.37&62.79&56.98 \\
    Cap+Know-LSTM&65.61&55.13&62.02&57.28 \\
    Att+Cap+Know-LSTM& \textbf{71.45}& \textbf{75.33}& \textbf{64.09}&\textbf{60.98} \\
     \Xhline{2\arrayrulewidth}
    \end{tabular}}
    \vspace{1pt}
    \caption{Toronto COCO-QA accuracy (\%) per category.}
    \vspace{-15pt}
    \label{tab3}  
\end{table}

Table \ref{tab2} reports the results. 
All of our proposed models outperform the \textbf{Baseline} and all of the competing state-of-the-art methods. Our final model \textbf{Att+Cap+Know-LSTM} achieves the best results. It surpasses the baseline by nearly 20\% and outperforms the previous state-of-the-art methods by around 15\%. \textbf{Att+Cap-LSTM} clearly improves the results over the \textbf{Att-LSTM} model. This proves that internal textual representation plays a significant role in the VQA task. The \textbf{Att+Know-LSTM} model does not perform as well as \textbf{Att+Cap-LSTM} , which suggests that the information extracted from captions is more valuable than that extracted from the KB.  \textbf{Cap+Know-LSTM} also performs better than \textbf{Att+Know-LSTM}. This is not surprising because the Toronto COCO-QA questions were generated automatically from the MS COCO captions, and thus the fact that they can be answered by training on the captions is to be expected.  This generation process also leads to questions which require little external information to answer. The comparison on the Toronto COCO-QA thus provides an important benchmark against related methods, but does not really test the ability of our method to incorporate extra information.  It is thus interesting that the additional external information provides any benefit at all.

\begin{table}[t!]
\centering
\resizebox{\linewidth}{!}{
 \begin{tabular}{lclccccc}
  \Xhline{2\arrayrulewidth}
  \multicolumn{1}{c}{}         & Our-Baseline &  & \multicolumn{5}{c}{Our Proposal}                                          \\ \cline{2-2} \cline{4-8}
  \multicolumn{1}{c}{Question} & VggNet       &  & Att              & Att+Cap          & Att+Know         &Cap+Know	& A+C+K            \\
  \multicolumn{1}{c}{Type}     & +            &  & +                & +                & +                &+		& +                \\
  \multicolumn{1}{c}{}         & LSTM         &  & LSTM             & LSTM             & LSTM             &LSTM		& LSTM             \\ \hline
  what is                      & 21.41        &  & 34.63            & 42.21            & 37.11            &35.58		& \textbf{42.52}   \\
  what colour                  & 29.96        &  & 39.07            & 48.65            & 39.68            &40.62		& \textbf{48.86}            \\
  what kind                    & 24.15        &  & 41.22            & 47.93            & 46.16            &44.04		& \textbf{48.05}   \\
  what are                     & 23.05        &  & 38.87            & 47.13            & 41.13            &39.73		& \textbf{47.21}            \\
  what type                    & 26.36        &  & 41.71            & 47.98            & 44.91            &44.95		& \textbf{48.11}           \\
  is the                       & 71.49        &  & 73.22            & 74.63            & 74.40            &73.78		& \textbf{74.70}   \\
  is this                      & 73.00        &  & 75.26            & 76.08            & \textbf{76.56}   &74.18		& 76.14            \\
  how many                     & 34.42        &  & 39.14            & 46.61            & 39.78            &44.20		& \textbf{47.38}   \\
  are                          & 73.51        &  & 75.14            & 76.01            & 75.75            &75.78		& \textbf{76.14}            \\
  does                         & 76.51        &  & 76.71            & 78.07            & 76.55            &77.17		& \textbf{78.11}   \\
  where                        & 10.54        &  & 21.42            & 25.92            & 24.13            &16.09		& \textbf{26.00}   \\
  is there                     & 86.66        &  & \textbf{87.10}            & 86.82            & 85.87            &85.26		& 87.01            \\
  why                          & 3.04         &  & 7.77             & 9.63             & 11.88            &9.99		& \textbf{13.53}            \\
  which                        & 31.28        &  & 36.60            & \textbf{39.55}   & 37.71            &37.86		& 38.70            \\
  do                           & 76.44        &  & 75.76            & 78.18            & 75.25            &74.91		& \textbf{78.42}            \\
  what does                    & 15.45        &  & 19.33            & 21.80            & 19.50            &19.04		& \textbf{22.16}            \\
  what time                    & 13.11        &  & 15.34            & 15.44            & \textbf{15.47}   &15.04		& 15.34            \\
  who                          & 17.07        &  & 22.56            & 25.71            & 21.23            &22.86		& \textbf{25.74}            \\
  what sport                   & 65.65        &  & 91.02            & 93.96            & 90.86            &91.75		& \textbf{94.20}   \\
  what animal                  & 27.77        &  & 61.39            & 70.65            & 63.91            &63.26		& \textbf{71.70}            \\
  what brand                   & 26.73        &  & 32.25            & 33.78            & 32.44            &31.30		& \textbf{34.60}            \\
  others                       & 44.37        &  & 50.23            & 53.29            & 52.11            &51.20		& \textbf{53.45}            \\ \hline
  Overall                      & 44.93        &  & 51.60            & 55.04            & 53.79            &52.31		& \textbf{55.96}            \\ \Xhline{2\arrayrulewidth}
 \end{tabular}}
\vspace{1pt}
\caption{Results on the open-answer task for various question types on VQA validation set. All results are in terms of the evaluation metric from the VQA evaluation tools. The overall accuracy for the model of \textbf{VggNet+ft+LSTM} is 50.01. Detailed results of different question types for this model are not shown in the table due to the limited space.}
\vspace{-13pt}
\label{tab4}
\end{table}

Table \ref{tab3} shows the per-category accuracy for different models. Surprisingly, the counting ability (see question type `Number') increases when both captions and external knowledge are included. This may be because some `counting' questions 
are not framed in terms of the labels used in the MS COCO captions.
Ren \etal also observed similar cases. In \cite{ren2015image} they mentioned that ``there was some observable counting ability in very clean images with a single object type but the ability was fairly weak when different object types are present''. We also find there is a slight increase for the `color' questions when the KB is used. Indeed, some questions like `What is the color of the stop sign?' can be answered directly from the KB, without the visual cue. %

\subsection{Results on the VQA}
Antol \etal in \cite{antol2015vqa} provide the VQA dataset which is intended to support ``free-form and open-ended Visual Question Answering''. They also provide a metric for measuring performance:
$\min\{\frac{\text{\# humans that said answer}}{3},1\}$
thus $100\%$ means that at least 3 of the 10 humans who answered the question  gave the same answer.
We have used the provided evaluation code %
to produce the results in Table~\ref{tab4}.

\begin{table*}[t]
\resizebox{\linewidth}{!}{
\begin{tabular}{lcccc}
 \multicolumn{2}{c}{\includegraphics[height=3cm]{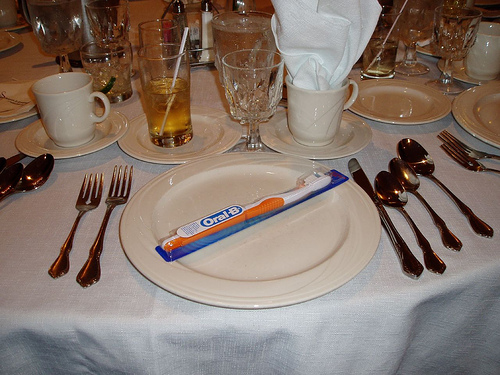}}&\includegraphics[height=3cm]{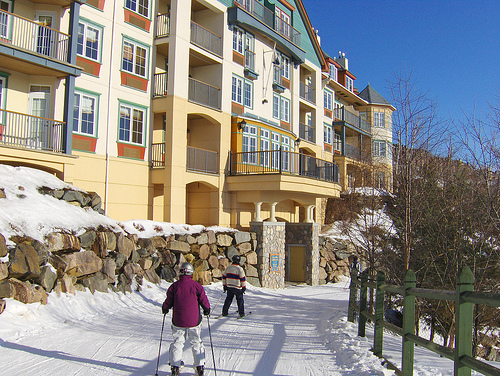}&\includegraphics[height=3cm]{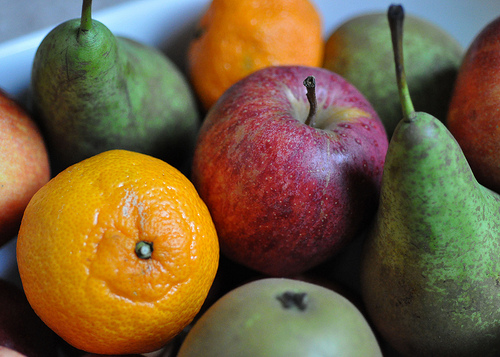}&\includegraphics[height=3cm]{img/results/008.jpg}\\
 \multicolumn{2}{c}{What color is the tablecloth?}&How many people in the photo?&What is the red fruit?&What are these people doing?\\ \hline
 \textit{Ours:} &\color{green}{white}&\color{green}{2}&\color{green}{apple}&\color{green}{eating}\\
 \textit{Vgg+LSTM:} &\color{red}{red}&\color{red}{1}&\color{red}{banana}&\color{red}{playing}\\
 \textit{Ground Truth:} &white&2&apple&eating\\ \Xhline{2\arrayrulewidth}
 \vspace{1pt}
\end{tabular}}
\resizebox{\linewidth}{!}{
\begin{tabular}{lcccc}
 \multicolumn{2}{c}{\includegraphics[height=3cm]{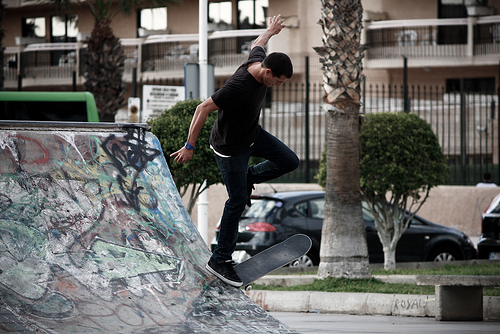}}&\includegraphics[height=3cm]{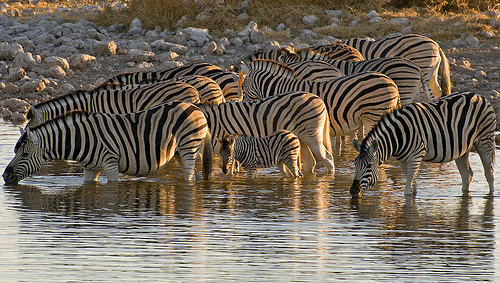}&\includegraphics[height=3cm]{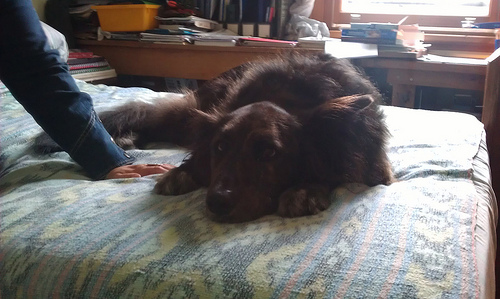}&\includegraphics[height=3cm]{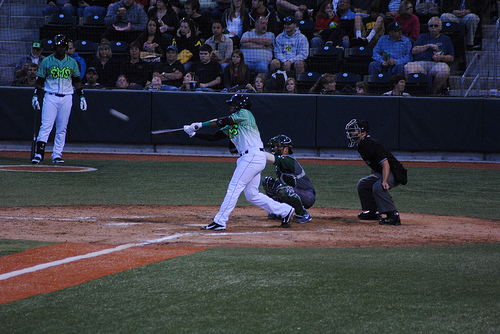}\\
 \multicolumn{2}{c}{Why are his hands outstretched?}&Why are the zebras in water?&Is the dog standing or laying down?&Which sport is this?\\ \hline
 \textit{Ours:} &\color{green}{balance}&\color{green}{drinking}&\color{green}{laying down}&\color{green}{baseball}\\
 \textit{Vgg+LSTM:} &\color{red}{play}&\color{red}{water}&\color{red}{sitting}&\color{red}{tennis}\\
 \textit{Ground Truth:} &balance&drinking&laying down&baseball\\ \Xhline{2\arrayrulewidth}
\end{tabular}}
\vspace{1pt} 
\caption{Some example cases where our final model gives the correct answer while the base line model \textbf{VggNet-LSTM} generates the wrong answer. All results are from VQA. More results can be found in the supplementary material.}
\label{results_examples}
\vspace{-2pt} 
\end{table*}

\vspace{-9pt}  
\paragraph{Evaluation}

Inspecting Table~\ref{tab4}, results on the VQA validation set, we see that the attribute-based \textbf{Att-LSTM} is a significant improvement over our \textbf{VggNet+LSTM} baseline. We also evaluate another baseline, the \textbf{VggNet+ft+LSTM}, which uses the penultimate layer of the attributes prediction CNN as the input to the LSTM. Its overall accuracy on the VQA is 50.01, which is still lower than our proposed models. Adding either image captions or external knowledge further improves the result. The model \textbf{Cap+Know} produces overall accuracy 52.31, slightly lower than Att+Know (53.79). This suggests that the attributes representation plays a more important role here.
Our final model \textbf{A+C+K-LSTM} produces the best results, outperforming the baseline \textbf{VggNet-LSTM} by 11\% overall. %

\begin{figure}[h!]
\vspace{-5pt}
\begin{center}
   \includegraphics[width=0.85\linewidth]{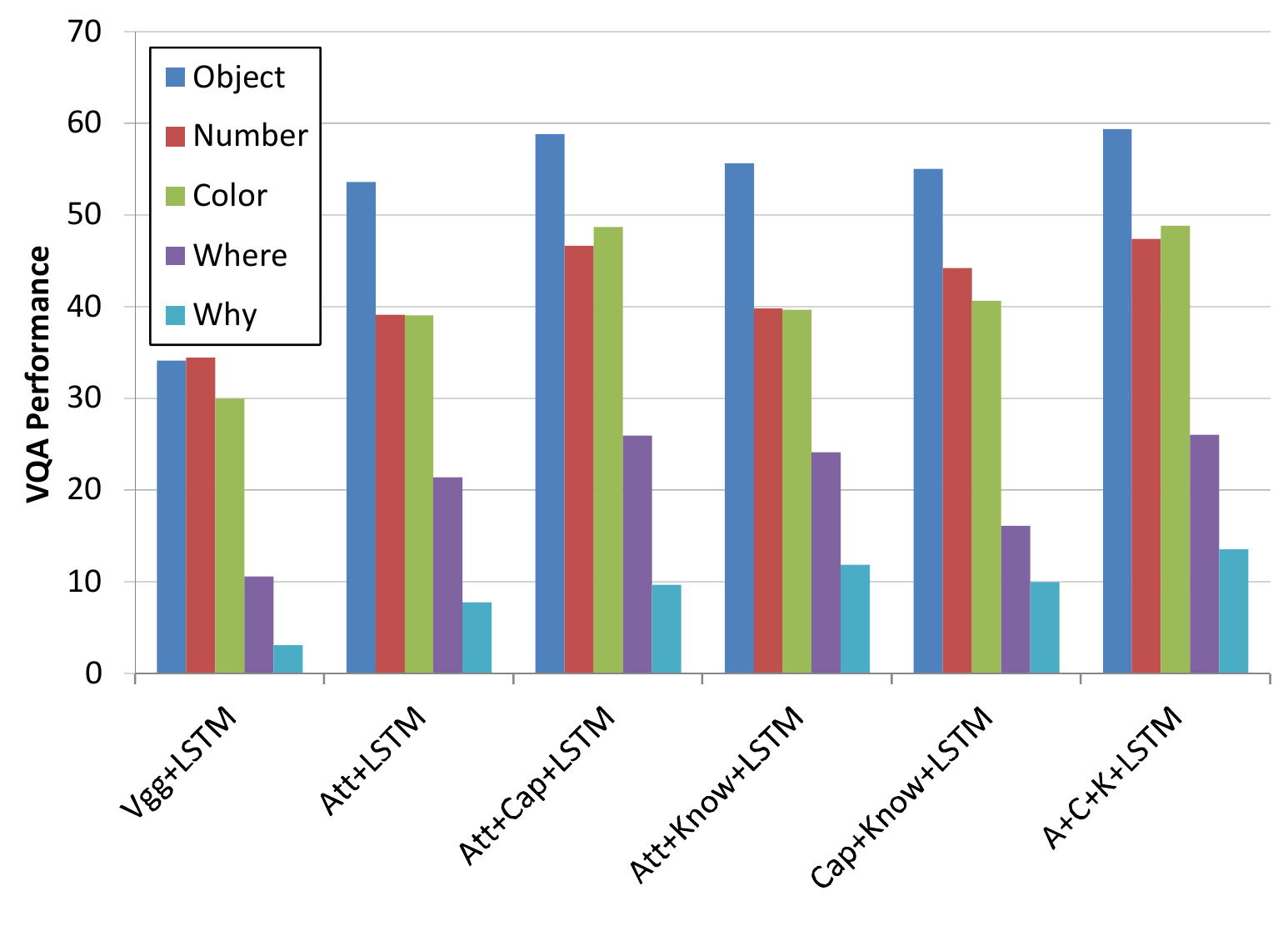}
\end{center}
\vspace{-13pt}
   \caption{Performance on five question categories for different models. The `Object' category is the average accuracy of question types starting with `what kind/type/sport/animal/brand...'.}
\vspace{-5pt}
\label{performance_trend}
\end{figure}

Figure~\ref{performance_trend} relates the performance of the various models on five categories of questions. The `object' category is the average of the accuracy of question types starting with `what kind/type/sport/animal/brand...', while the `number' and `color' category corresponds to the question type `how many' and `what color'. 
The performance comparison across categories is of particular interest here because answering different classes of questions requires different amounts of external knowledge.  The `Where' questions, for instance, require knowledge of potential locations, and `Why' questions typically require general knowledge about people's motivation. `Number' and `Color' questions, in contrast, can be answered directly. The results show that for `Why' questions, adding the KB improves performance by more than $50\%$ (Att-LSTM achieves $7.77\%$ while Att+Know-LSTM achieves $11.88\%$), and that the combined A+C+K-LSTM achieves $13.53\%$.

Table \ref{results_examples} provides some indicative results, more results can be found in the supplementary material, including failure cases.

We have also tested on the VQA test-dev  
and test-standard\footnote{http://www.visualqa.org/challenge.html} 
consisting of 60,864 and 244,302 questions (for which ground truth answers are not published) using our final A+C+K-LSTM model, and evaluated them on the VQA evaluation server. %
Table \ref{server_results} shows the server reported results.

\begin{table}[t!]
\centering
\resizebox{\linewidth}{!}{
\begin{tabular}{ccccccccc}
\Xhline{2\arrayrulewidth}
\multicolumn{1}{c|}{}         & \multicolumn{4}{c|}{Test-dev}                       & \multicolumn{4}{c}{Test-standard} \\
\multicolumn{1}{c|}{}         & All   & Y/N   & Num   & \multicolumn{1}{c|}{Others} & All    & Y/N    & Num    & Others \\ \hline
\multicolumn{1}{c|}{Question \cite{antol2015vqa}}   & 40.09 & 75.66 & 36.70 & \multicolumn{1}{c|}{27.14}  & -  & -  & -  & -  \\
\multicolumn{1}{c|}{Image \cite{antol2015vqa}}   & 28.13 & 64.01 & 0.42 & \multicolumn{1}{c|}{3.77}  & -  & -  & -  & -  \\
\multicolumn{1}{c|}{Q+I \cite{antol2015vqa}}   & 52.64 & 75.55 & 33.67 & \multicolumn{1}{c|}{37.37}  & -  & -  & -  & -  \\
\multicolumn{1}{c|}{LSTM Q \cite{antol2015vqa}}   & 48.76 & 78.20 & 35.68 & \multicolumn{1}{c|}{26.59}  & 48.89  & 78.12  & 34.94  & 26.99  \\
\multicolumn{1}{c|}{LSTM Q+I \cite{antol2015vqa}} & 53.74 & 78.94 & 35.24 & \multicolumn{1}{c|}{36.42}  & 54.06  & 79.01  & 35.55  & 36.80  \\
\multicolumn{1}{c|}{Human \cite{antol2015vqa}}    & -     & -     & -     & \multicolumn{1}{c|}{-}      & 83.30  & 95.77  & 83.39  & 72.67  \\ \hline
\multicolumn{1}{c|}{Ours}      & \textbf{59.17} & \textbf{81.01} & \textbf{38.42} & \multicolumn{1}{c|}{\textbf{45.23}}  & \textbf{59.44}  & \textbf{81.07}  & \textbf{37.12} & \textbf{45.83}  \\
\Xhline{2\arrayrulewidth}
\end{tabular}}
\vspace{1pt}
\caption{VQA Open-Ended evaluation server results for our method. Accuracies for different answer types and overall performances on test-dev and test-standard datasets are shown.}
\vspace{-8pt}
\label{server_results}
\end{table}

Antol \etal \cite{antol2015vqa} provide several results for this dataset. In each case they encode the image with the final hidden layer from VggNet, and questions are encoded using a BOW representation. A softmax neural network classifier with 2 hidden layers and 1000 hidden units (dropout 0.5) in each layer with tanh non-linearity is then trained, the output space of which is the 1000 most frequent answers in the training set. They also provide an LSTM model followed by a softmax layer to generate the answer. Two versions of this approach are used, one which is given both the question and the image, and one which is given only the question (see~\cite{antol2015vqa} for details). %
Our final model outperforms the listed approaches according to the overall accuracy and all answer types. %

%
\section{Conclusion}
\label{Sec:Conclusion}
\vspace{-3pt}
Open-ended visual question answering is an interesting challenge for computer vision, not least because it represents a move away from purely image-based analysis and towards a broader form of artificial visual intelligence.  In this paper we have shown that it is possible to extend the state-of-the-art RNN-based VQA approach so as to incorporate the large volumes of information required to answer general, open-ended, questions about images.  The knowledge bases which are currently available do not contain much of the information which would be beneficial to this process, but none-the-less can still be used to significantly improve performance on questions requiring external knowledge (such as `Why' questions).  The approach that we propose is very general, however, and will be applicable to more informative knowledge bases should they become available. At the time of writing this paper, our system performs the best on two large-scale VQA datasets and produces promising results on the VQA evaluation server. 
Further work includes generating knowledge-base queries which reflect the content of the question and the image, in order to extract more specifically related information. 

It seems inevitable that successful general visual question answering will demand access to a large external knowledge base.  We have shown that this is possible using the LSTM-based approach, and that it improves performance.  We hope this might lead the way to a visual question answering approach which is capable of deeper analysis of image content, and even common sense.

{\bf Acknowledgements}
This research was in part supported by the Data to Decisions Cooperative Research Centre.

{\small
\bibliographystyle{ieee}
\bibliography{myref}
}

\end{document}